\documentclass[english]{article}
\usepackage[latin9]{inputenc}
\usepackage{amsmath}
\usepackage{amssymb}
\usepackage{graphicx}

\makeatletter

\providecommand{\tabularnewline}{\\}

\@ifundefined{showcaptionsetup}{}{%
 \PassOptionsToPackage{caption=false}{subfig}}
\usepackage{subfig}
\makeatother

\usepackage{babel}
\usepackage{listings}

\begin{document}

\title{The Computational Theory of Intelligence: Applications to Genetic
Programming and Turing Machines}

\author{Daniel Kovach}
\maketitle
\begin{abstract}
In this paper, we continue the efforts of the Computational Theory
of Intelligence (CTI) by extending concepts to include computational
processes in terms of Genetic Algorithms (GA's) and Turing Machines
(TM's). Active, Passive, and Hybrid Computational Intelligence processes
are also introduced and discussed. We consider the ramifications of
the assumptions of CTI with regard to the qualities of reproduction
and virility. Applications to Biology, Computer Science and Cyber
Security are also discussed.
\end{abstract}

\section{Introduction}

In this work, we carry on the ideas proposed in \cite{cti1,cti2}.
In \cite{cti1}, we posited some very basic assumptions as to what
intelligence means as a mapping. Using our analysis, we formulated
some axioms based on entropic analysis. In the paper that followed
\cite{cti2} we talked about how ideas or memes can aggregate together
to form more complex constructs.

Here, we extend the concept of intelligence as a computational process
to computational processes themselves, in the classic Turing sense.

The structure of this paper is as follows. In section \ref{sec:Passive,-Active,-and},
we provide deeper insights as to different types of computational
intelligence, specifically in the light of executable code, section
\ref{sec:Reproduction} the qualities of reproduction, virility, and
nested programs, section \ref{sec:Experimentation} presents some
experimental results, and sections \ref{sec:Applications-to-Computional}
and \ref{sec:Applications-to-Cyber} discuss applications. In the
conclusion, we recapitulate and discuss future work.

\section{Extending Concepts\label{sec:Passive,-Active,-and}}

Consider the definition of computational intelligence from \cite{cti1}.
Given the two sets $\mathbb{S}$ and $\mathbb{O}$, which represent
input and output sets, respectively, we introduce the mapping 

\begin{equation}
\mathbb{I}:\mathbb{S}\rightarrow\mathbb{O},\label{eq:I}
\end{equation}

where $\mathbb{I}$ essentially makes a prediction based on the input.
The disparity between a prediction and reality can then be used to
update $\mathbb{I}$ in such a way that it can climb the gradient
of the learning function $\mathbb{L}$ so as to improve future performance.
Although we touched upon the application of this framework to genetic
algorithms \cite{cti1}, it will be insightful to revisit it in more
detail.

\subsection{Active, Passive, and Hybrid Computational Intelligence\label{sub:Active,-Passive,-and}}

Let us clarify the terminology. By \emph{genetic algorithm} (GA),
we mean a collection or \emph{population} $\Pi=\{\pi_{i}\}$ of potential
solutions to a particular task, which evolves as time progresses according
to some fitness function, $\mathbb{F}$. The function $\mathbb{F}$
is synonymous with learning function $\mathbb{L}$ from our previous
formulation. It evaluates the quality of $\Pi$ as a solution to the
task. Specifically, if our task or \emph{program }$P$ is a function
of $\Pi$, then the fitness function $\mathbb{F}$ is nothing more
than

\begin{equation}
\mathbb{F}=\mathbb{F}(P(\Pi)).
\end{equation}

Note that the population $\Pi$ is not only our input set $\mathbb{S}$
but our output set $\mathbb{O}$ in the representation \ref{eq:I}.
Furthermore, the mapping $\mathbb{I}$ is nothing but a collection
of mutation operations to update the $\Pi$ at iteration $t$ to iteration
$t+1$. That is

\begin{equation}
I:\Pi_{t}\rightarrow\Pi_{t+1}.\label{eq:intelligence mapping pis}
\end{equation}

Some common mutation operations of genetic algorithms are \cite{bio inspired AI}:
\begin{enumerate}
\item Reproduction: selecting individuals within a population and copying
them without alteration into the new population.
\item Crossover: produces new offspring from elements of two other \emph{parents.}
\item Mutation: one of the central concepts of genetic algorithms. Mutation
perturbs the population or its constituents in some way. This includes
altering the elements of the population directly, or changing their
placement in the population in some way such as swapping their indices.
\item Editing: a manner in which more complicated functionality can be reduced
to simpler algebraic rules.
\item Adding/Deleting: the ability to add a new element to a population
or remove an existing element from the population.
\item Encapsulating: the process of creating a new element which represents
an aggregate of functionality from one or more other elements in the
population.
\end{enumerate}
Some consideration should be given to the statement

\begin{equation}
\mathbb{I}_{t+1}=\mathbb{I}_{t}+\nabla\mathbb{F}_{t}\label{eq:intelligence mapping update}
\end{equation}

from \cite{cti1}. For this equation to make sense, not only must
the algebraic operation '+' be meaningfully defined, but the gradient
operator implies that $\mathbb{F}$ is differentiable. The gradient
operation makes sense only if the domain of $\mathbb{F}$ is a Banach
space . In this context, we cannot make such a claim. Thus equation
\ref{eq:intelligence mapping update} becomes 

\begin{equation}
\mathbb{I}_{t+1}=\mathbb{I}_{t}+\Delta\mathbb{F},\label{eq:intelligence mapping update-1}
\end{equation}

where $\Delta\mathbb{F}=\mathbb{F}(\Pi_{t})-\mathbb{F}(\Pi_{t-1})$.
Observe that although the mathematical formulation is different, the
intent still holds. The '+' operator communicates the number of permutations
proportional to $\Delta\mathbb{F}$. 

Since the information is retained in the population, and only improved
through evolutionary perturbations, we consider this an example of
\emph{passive computational intelligence}. This stands in contrast
to \emph{active computational intelligence}, where we are updating
the mapping $\mathbb{I}$ itself. Any combination of the two methods
can be considered a form of \emph{hybrid computational intelligence}.

\subsection{Genetic Algorithms}

Recall the entropic considerations from \cite{cti1}, where we specified
that
\begin{itemize}
\item \emph{Computational intelligence is a process that locally minimizes
and globally maximizes entropy.}
\end{itemize}
Since the population $\Pi$ is the output set, it is the entropy of
the population that must be minimized by the intelligence process.
Equivalently, we can say that as the amount of iterations becomes
sufficiently large, the difference between $\Pi_{t}$ and $\Pi_{t-1}$
becomes increasingly negligible. In other words, assuming the subtraction
operation is meaningfully defined, we have

\begin{equation}
\lim_{t\rightarrow\infty}\Pi_{t+1}-\Pi_{t}=0.\label{eq:difference in population}
\end{equation}

This makes sense, for we expect that as time increases, the fitness
of our solution reaches an optimum, so that eventually there is no
need to update. In other words, our population becomes more consistent
over time.

Of course, the typical considerations regarding genetic algorithms
such the necessity of avoiding local optima, the proper choice of
a fitness function, the complexity of the algorithm itself, among
others, still hold \cite{bio inspired AI}.

As for global entropy considerations, we must consider the role of
the population in its realization under its program $P$. As mentioned
before, the total entropy of a system can be written as:

\begin{equation}
S=\sum_{i=1}^{N}s_{i}
\end{equation}

where $\{s_{i}\}$ represent each respective source of entropy. In
particular, under the considerations of this section, we have

\begin{equation}
S=s_{\Pi}+s_{P},\label{program entropy}
\end{equation}

where $s_{\Pi}$ and $s_{P}$ represent the entropy due to the population
itself and the entropy from the program $P$, respectively.

To calculate the entropy, we used the Renyi entropy of order $\alpha$,
where $\alpha\geq0$ and $\alpha\neq1$ defined as

\begin{equation}
\mathbb{H_{\alpha}}(X)=\frac{1}{1-\alpha}\log\left[\sum_{i=1}^{N}\mathbb{P}\left[x_{i}\right]^{\alpha}\right].\label{eq:renyi}
\end{equation}

where $\mathbb{P}[x_{i}]$ is the probability associated with state
$x_{i}\in X$.

\subsection{Turing Machines\label{sec:Turing-Tape}}

Let us consider the discussions of the previous sections in terms
of code that is executable in the Turing sense. We will apply the
formulations we have considered thus far and make improvements to
refine the theory in this light.

Recall that a Turing machine $M$ is a hypothetical device that manipulates
symbols on a tape, called a Turing tape, according to a set of rules.
Despite its simplicity, the Turing machine analogy can be used to
describe certain aspects of modern computers \cite{turing and computers},
and more \cite{dna turing complete}.

The Turing tape has some important characteristics. The tape contents,
called the \emph{alphabet}, encode a given table of rules. Using these
rules the machine manipulates the tape. It can write and read to and
from the tape, change the configuration, or \emph{state }$\Sigma$
of the machine, and move the tape back and forth \cite{turing machines}.

Typically, the instructions on the Turing tape consist of at least
one instruction that tells the machine to begin computation, or a
START instruction, and one to cease computation, called a STOP or
HALT instruction \cite{turing machines}. Other instructions might
include MOVE, COPY, etc, or even instructions that encode no operation
at all. Such instructions are commonly called NOOP's (NO OPeration).
The total number of instructions depends on the situation and the
application at hand, but for the purposes of this paper, we will consider
only tape of finite length for which the instruction set is also finite
and contains at least the START and STOP instructions. In fact, for
our purposes, the minimum amount of opcodes a program needs to run,
or to be \emph{executable,} are the START and STOP instructions.

Now, consider the set of all instructions available to a particular
$M$. To be consistent with our notation thus far, we will denote
this set as $\widetilde{\pi}$, where

\begin{equation}
\widetilde{\pi}=\{\pi_{0}=\mathrm{START},...,\pi_{n}=\mathrm{STOP}\}
\end{equation}

From this set, a particular sequence of instructions $\Pi$ can be
formed. Although many synonymous terms abound for this sequence of
instructions, including a code segment, executable, or tape, we will
refer to it simply as the \emph{code}. We consider the execution or
implementation $\Psi$ to be the execution of the code on the Turing
machine $M$, which produces could produce another code segment, and
change the machine state. Expressed in notation:

\begin{equation}
\Psi:\widetilde{\Pi}\times\widetilde{\Sigma}\rightarrow\widetilde{\Pi}\times\widetilde{\Sigma}.\label{eq:code implementation}
\end{equation}

where $\widetilde{\Pi}$ and $\widetilde{\Sigma}$ represent the set
of all code and machine states, respectively. For the majority of
this paper, it is the code that is of primary concern, and thus we
can omit the machine state to simplify the notation when this context
is understood:

\begin{equation}
\Psi:\widetilde{\Pi}\rightarrow\widetilde{\Pi}
\end{equation}

Our fitness function $\mathbb{F}$ is a measure the efficaciousness
of the implementation of our program by some predefined standard.
Later, we will present some specific examples used in application.

We are still free to think of the intelligence mapping $\mathbb{I}$
as a perturbation mapping as we did in section \ref{sub:Active,-Passive,-and}.
The only difference is the physical interpretation of the term 'population'.
This case concerns machine code instructions. As with genetic algorithms,
perturbing code may advance or impede the fitness of the program.
In fact, it may even destroy the ability to execute at all, if one
either the START or STOP operations are removed.

A final note should be made as to the entropic considerations of the
Turing Machine. In this light, we must take into account all the workings
of the machine and thus \ref{program entropy} becomes

\begin{equation}
S=s_{\Pi}+s_{M}\label{eq:entropy machine}
\end{equation}

where $s_{M}$ is the entropy incurred by the Turing Machine, which
includes its state $\Sigma$.

\section{Reproduction, Viruses, and Nested Programs\label{sec:Reproduction}}

In this section, we will continue the above analysis, and discussed
some qualities that emerge based on our assumption.

\subsection{Formalisms of Code and Other Qualities}

We can apply set theoretic formalisms in the usual sense. For example,
consider a code $\Pi$ and an instruction, $\pi\in\widetilde{\pi}$.
Then the notation $\pi\in\Pi$ communicates the fact that $\pi$ is
an instruction in $\Pi$. Further, set theoretic operations follow
naturally. Let $\Pi_{1},\Pi_{2}\in\widetilde{\Pi}$. In the case of
the union operation, we have

\begin{equation}
\Pi_{1}\cup\Pi_{2}=\{\Pi_{1},\Pi_{2}\}
\end{equation}

which is a set containing the two programs $\Pi_{1}$ and $\Pi_{2}$.
Similarly, with the intersection operation, we have

\begin{equation}
c\in\Pi_{1}\cap\Pi_{2}\Leftrightarrow c\in\Pi_{1}\wedge c\in\Pi_{2},
\end{equation}

or the set of all contiguous code segments $c=\{\pi\}$ that are contained
by both $\Pi_{1}$ and $\Pi_{2}$. Note that this code segment need
not be executable. Also, if it is of particular utility in the program
it may be \emph{degenerate} in that it may appear multiple times throughout
the code.

For set inclusion, we say that

\begin{equation}
\Pi_{1}\subset\Pi_{2}\Leftrightarrow\pi\in\Pi_{2}\forall\pi\in\Pi_{1}
\end{equation}

Proper containment evolves naturally if $\Pi_{1}=\Pi_{2}$.

\subsection{Metrics and Neighborhoods}

The differences between code and real numbers or sets in conventional
mathematics make it at first difficult to form a distance metric $d$
on them. Still, some have been explored such as the Levenshtein distance
\cite{levenshtein}, Damerau-Levenshtein distance \cite{demerau levenshtein},
Hamming distance \cite{hamming}, and the Jaro-Winkler distance \cite{Jaro Winkler}.

Given a metric $d$, we can form an open ball \cite{real analysis},
or neighborhood $B$ centered at some code $\Pi_{0}$ consisting of
all programs $\Pi$ such that for some real number $\epsilon$,$d(\Pi_{0},\Pi)<\epsilon$.
In other words

\begin{equation}
B(\Pi_{0},\epsilon)=\{\Pi:d(\Pi_{0},\Pi)<\epsilon\}.
\end{equation}

We can utilize the developments above to relax the definition of set
inclusion with respect to the metric, $d$. Specifically, define

\[
\Pi_{1}\overset{\epsilon}{\subset}\Pi_{2}\Leftrightarrow d(\Pi_{1},\Pi)<\epsilon,\Pi\subset\Pi_{2}
\]

Of course, similar considerations can be extended to proper containment.
Such code segments that lie suitably close to each other with respect
to $d$ will be referred to as \emph{polymorphic }with respect to
$d$.

\subsection{Reproduction}

Consider the mapping

\begin{equation}
\Psi_{i}(\Pi)=\Pi'
\end{equation}

where $\Pi'$ is in the neighborhood of $\Pi$. Of particular interest
is when $\Pi'$ not only lies within the neighborhood of $\Pi$ but
produces a machine state $\Sigma'$ suitably close to that of $\Pi$,
as valued by some meaningful metric. We note in particular the condition
when $\Pi'$ also encodes for reproduction. Unless in the presence
of polymorphic programs \cite{polymorphic}, we will assume the code
copies itself exactly, and replace $\Pi'$ with $\Pi$ in notation.

Sequential program execution will yield the same result

\begin{equation}
\Psi_{t}=\Psi(\Psi_{t-1}),\Psi_{0}=\Psi(\Pi)
\end{equation}

where

\begin{equation}
\Psi_{i}(\Pi)=\Pi,\forall i.
\end{equation}

Suppose that a particular tape is able to produce, say $N$ copies
of itself. Then the entropy of the system is simply

\begin{equation}
S=s_{M}+\sum_{i=0}^{N}s_{\Pi_{i}},
\end{equation}

where $s_{0}$ refers to the original copy of the tape. By comparison
with equation \ref{eq:entropy machine}, the entropy is proportional
to the number of progeny in addition to the machine state.

\subsection{Viruses\label{sub:Viruses}}

Thus far, we have been considering only predefined programs. But a
well known concept from nature concerns when external code is injected
into a host program, or otherwise hijacks the computational resources
of the system. One key characteristic of viruses is their ability
to proliferate. Consider $\nu$ such that

\begin{equation}
\nu\in\Psi(\Pi),\nu\subseteq\Pi
\end{equation}

If the viral code, $\nu$ is executable, we will call the virus $\nu$-executable.
If the viral code also can reproduce, that is $\nu\in\Psi(\nu)$,
we will call this quality $\nu$-reproductive. Finally, it is often
the case that viral code is not reproductive or executable. Its task
is to merely copy or install some payload \cite{payload}, here denoted
by $\rho$. In this case,

\begin{equation}
\rho\in\Psi(\Pi),\rho\subseteq\nu\subseteq\Pi
\end{equation}

Viral code most often is pernicious, though this is not necessarily
the case. For example, up to 8\% of the human genome derives from
retroviruses \cite{retro viruses}. To determine the nature of the
effect of the virus, we look at the change in fitness before and after
its influence.

The influence of viral code on program fitness falls into three categories.
Consider $\Pi,\nu\bigcap\Pi=\emptyset$, and $\Pi',\nu\bigcap\Pi'=\nu,\Pi=\Pi'-\nu$.
In other words, $\Pi'$ code in the presence of a virus, from the
otherwise pristine $\Pi$. The three cases are $\mathbb{F}(\Pi')-\mathbb{F}(\Pi)>0$,
$\mathbb{F}(\Pi')-\mathbb{F}(\Pi)=0$, and $\mathbb{F}(\Pi')-\mathbb{F}(\Pi)<0$.
These cases are of enough significance that they deserve their own
nomenclature. We will refer to these cases as commensalistic, symbiotic,
and parasitic, respectively, out of consistency with their counterparts
from biological sciences.

\subsection{Nested Programs}

One final case to be considered is when a program or multiple programs
can contribute to produce aggregate programs of greater complexity.
In such a case, the program, when executed, would not only alter machine
state and produce code as in \ref{eq:code implementation} but will
also contain a product code segment, and product code machine state, 

\begin{equation}
\Psi:\widetilde{\Pi}\times\widetilde{\Sigma}\rightarrow\widetilde{\Pi}\times\widetilde{\Sigma}\times\widetilde{\Pi}^{1}\times\widetilde{\Sigma}^{1}\label{eq: base nested phi}
\end{equation}

Introduce the program

\begin{equation}
\Psi^{1}=\widetilde{\Pi}^{1}\times\widetilde{\Sigma}^{1}\rightarrow\widetilde{\Pi}^{1}\times\widetilde{\Sigma}^{1}
\end{equation}

Of course, we could nest this process and define

\begin{equation}
\Psi^{1}=\widetilde{\Pi}^{1}\times\widetilde{\Sigma}^{1}\rightarrow\widetilde{\Pi}^{1}\times\widetilde{\Sigma}^{1}\times\widetilde{\Pi}^{2}\times\widetilde{\Sigma}^{2},
\end{equation}

with $\Psi$ from \ref{eq: base nested phi} denoted as $\Psi^{0}$,
we could write

\begin{equation}
\Psi^{n}[\Psi^{n-1}[...\Psi^{0}[\Pi^{0}\times\Sigma^{0}]...]]\rightarrow\widetilde{\Pi}^{n}\times\widetilde{\Sigma}^{n}
\end{equation}

where the superscript indicates the level of nesting of the program.

Just as we discussed programs producing multiple progeny, we can also
let them produce multiple product code. We will denote each copy with
a subscript $i_{m}=\{0,1,...,I_{m}\}$. Thus the above becomes

\begin{equation}
\Psi^{n}[\Psi_{i_{n-1}}^{n-1}[...\Psi_{i_{0}}^{0}[\Pi_{i_{0}}^{0}\times\Sigma_{i_{0}}^{0}]...]]\rightarrow\widetilde{\Pi}_{i_{n}}^{n}\times\widetilde{\Sigma}_{i_{n}}^{n}.
\end{equation}

To determine the total entropy, simply sum over all product code.

\begin{equation}
S=s_{M}+\sum_{j=0}^{I_{k}}\sum_{k=0}^{n}\widetilde{\Pi}_{j}^{k}\times\widetilde{\Sigma}_{j}^{k}
\end{equation}

This time we see that the entropy of the code varies exponentially
with product and copy code.

\section{Experimentation\label{sec:Experimentation}}

Consider the following scenario. We have a set of instructions, and
a construct for a tape that encodes them. In the following experiments,
the code is created randomly. We apply a GA to the code to acheive
interesting results. The code involved had no write back capabilities,
and thus the experiment was purely passive.

\subsection{Genetics and Codons}

A major focus for future work is the application of this research
to computational molecular biology. Hence, we chose to implement the
instruction set that nature has chosen for genetics in the following
experiments. 

For the base instruction alphabet, we will choose that of RNA \cite{genetics}.
This instruction set is composed of the 'bits' Adenine (A), Uracil
(U), Guanine (G), and Thymine (T). Each instruction can be represented
as three of these quaternary bits, for a total of 64 possible outcomes.
We have chosen this quaternary codon formulation to be the basis of
our tape.

In both experiments, the choice of operational codes, or \emph{opcodes,}
are taken directly from these 64 quaternary bit codons. The choice
to represent a opcode by a particular codon is completely arbitrary.
All codons appear with the same probability in experiment, with the
exception of the STOP opcode, as we will see in section \ref{sub:First-Instruction-Set}.

We present two instructions sets, and two different experiments. In
the first experiment, we determine how many iterations our genetic
algorithm takes to produce code that is executable, and code that
reproduces itself. In the second experiment, we compare the amount
of reproductions of code with the total entropy incurred by the simulation.

\subsection{First Instruction Set\label{sub:First-Instruction-Set}}

For the first experiment we constructed a robust instruction set consisting
of the following:
\begin{enumerate}
\item START (AAA): Commences program execution.
\item STOP (AUA, ATC, ATG): Halts program execution. Observe that there
are three STOP codons, but only one START codon. This mirrors the
amount of codons in RNA and DNA observed in nature \cite{three stop codons}.
\item BUILD\_FR (CUC): Copies to product code starting from the next respective
instruction to that which comes before the BUILD\_TO (GCG) instruction.
\item COND (UUC, UUA, GAA): Sets an internal variable in the Turing machine
state called a \emph{flag}. A flag can be thought of as an internal
Boolean variable. Every time one of the COND opcodes is encountered
the sign of the flag is switched, that is $\mathrm{COND}=\neg\mathrm{COND}$.
\item IF (AAU): Only executes the following instruction if the COND flag
has been set.
\item COPY\_ALL (AAG): Copies the entire tape to progeny.
\item COPY\_FR (CCC): Copies to progeny code starting from the next respective
instruction to that which comes before the COPY\_TO (GGG) instruction.
\item JUMP\_FAR\_FR, (CUU) and JUMP\_NEAR\_FR, (AGA): both require a JUMP\_TO,
(CAC, GUG) instruction. The JUMP family of instructions is designed
to continue program execution at the index of the JUMP\_TO instruction.
The former will find the JUMP\_TO address that is farthest away, while
the JUMP\_NEAR finds the closest JUMP\_TO instructions. If there is
no multiplicity in the JUMP\_TO instruction, then JUMP\_FAR\_FR, (CUU)
and JUMP\_NEAR\_FR will produce the same result.
\item REM\_FR, (GCU) and REM\_TO, (UAA): Removes all code in between these
instructions.
\end{enumerate}

\subsection{Second Instruction Set}

Although the code of the previous section was robust, it was redundant.
For example, the functionality of the code to copy itself to progeny
should be a desireable trait of the \emph{code itself}, not just an
opcode that accomplishes the task in one step. Further, the remove
operations can be handled via mutation operations (albeit far less
efficiently) in the genetic algorithm itself. 

Note that in Experiment 1, most operations have a partner or \emph{conjugate}
opcode. The necessity of conjugate pairs of instructions follows from
the fact that two pieces of information are required to complete the
instruction, as shown below:

\begin{lstlisting}
[instruction]...<code>...[instruction dual]
\end{lstlisting}

We will call such instructions \emph{dual}. The necessity for a predefined
conjugate opcode can be mitigated by using the concept of \emph{addressing}.
Instead of conjugate opcodes, the instruction immediately following
the dual can be thought of as an argument. The machine uses this address
(unless the address is represents opcode) as a manner of locating
the conjugate to the dual and thus reducing the instruction set. Specifically:

\begin{lstlisting}
[instruction][address]...<code>...[address]
\end{lstlisting}

Under these considerations (and omitting the ability to build to a
product), we have need for only six instructions: START, STOP, IF,
COND, COPY, and JUMP. The first four instructions are exactly the
same as in Experiment 1. The COPY instruction concatenates section
of the tape to progeny, and the JUMP instruction continues program
execution at a specified address. These final two are conjugates,
the duals of which are handled via the addressing method entailed
above. Notice although the lexicon is different, they have the same
functionality as their respective counterparts in the first instruction
set.

\subsection{Experiment 1: Execution and Reproduction}

The goal of the first instruction set was to produce code that was
executable, and code that could reproduce. The code was perturbed
by a GA randomly. That is, no fitness function was used in the perturbation
of the code strings. Each epoch of the simulation terminated once
an executable or reproducable code was discovered. For each experiment,
a maximum of 1,000,000 computational iterations were allowed.

The results of the first experiment using the first instruction set
are shown in table \ref{tab:Results:-Experiment-1}. The results of
the first experiment using the second instruction set are summarized
in table \ref{tab:Results:-Experiment-2}. Observe that the number
iterations in the first instruction set was much less than that of
the second, which is likely due to the disparity in robustness between
the two. Also note how wildly the results vary especially in terms
of the reproductive capabilities of the second instruction set as
demonstrated by the large standard deviation in results.

\begin{table}
\centering{}\protect\caption{\label{tab:Results:-Experiment-1}Results: Experiment 1, Instruction
Set 1}
\begin{tabular}{|c|c|c|c|}
\hline 
Experiment & Average & Std. Dev & Experiments\tabularnewline
\hline 
\hline 
Executable Code & 184.083 & 126.202 & 100,000\tabularnewline
\hline 
Reproductive Code & 330.964 & 243.924 & 96,964\tabularnewline
\hline 
\end{tabular}
\end{table}

\begin{table}
\begin{centering}
\protect\caption{\label{tab:Results:-Experiment-2}Results: Experiment 1, Instruction
Set 2}

\par\end{centering}

\centering{}%
\begin{tabular}{|c|c|c|c|}
\hline 
Experiment & Average & Std. Dev & Experiments\tabularnewline
\hline 
\hline 
Executable Code & 283.387 & 210.359 & 1,000,000\tabularnewline
\hline 
Reproductive Code & 2003.24 & 96504.8 & 1,000,000\tabularnewline
\hline 
\end{tabular}
\end{table}

\subsection{Experiment 2: Reproduction and Entropy}

Upon answering the basic questions like the amount of computational
iterations necessary to produce executable and reproductive code,
the next experiment focused on the total entropy incurred by a given
code segment and its offspring. Here, recall that we are considering
the entropy produced by not only a given tape, but its progeny as
well. We will continue with the alphabets of Experiment 1 and 2, respectively.

We bounded the total computational iterations per experimental run
at 1,000,000. The maximum progeny allowed was 50. Mutations were applied
by the GA according to a fitness function which evaluated the quality
of the code with respect to its Renyi Entropy (of order $\alpha=2$).

\begin{figure}
\subfloat[Reproduction vs. Entropy (Alphabet 1)]{

\includegraphics[scale=0.25]{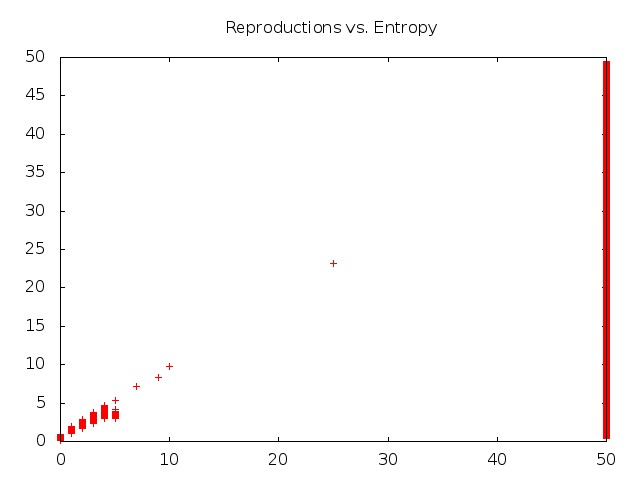}}\hfill{}\subfloat[Reproduction vs. Entropy (Alphabet 2)]{\includegraphics[scale=0.25]{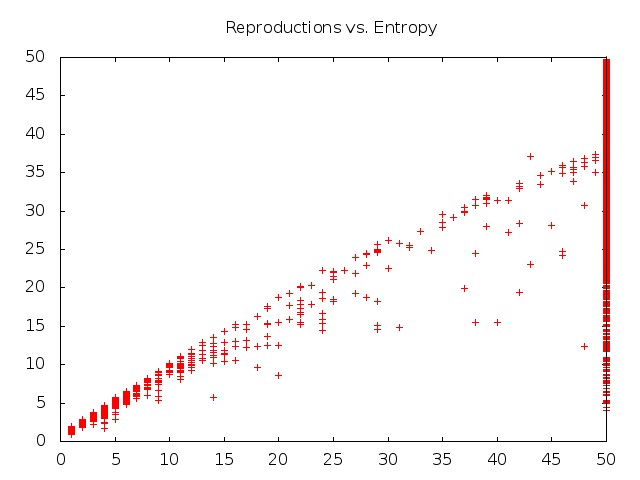}}

\protect\caption{Graphs of reproductive iterations plotted against the total entropy
of the simulation, for the first and second instruction sets, respectively.\label{fig:reproduction vs entropy}}
\end{figure}

The results are summarized in table \ref{tab:Results:-Reproduction-vs.}.
Observe that the entropy and the amount of reproductions appear to
be correlated as demonstrated by the r-values in table \ref{tab:Results:-Reproduction-vs.}
and in the correlation maps in Figure \ref{fig:reproduction vs entropy}.
Note that some programs appear to have acheived reproductive qualities
without substantially increasing entropy but the converse is not true.

\begin{figure}
\subfloat[Program Behavior from Sample 457]{

\includegraphics[scale=0.25]{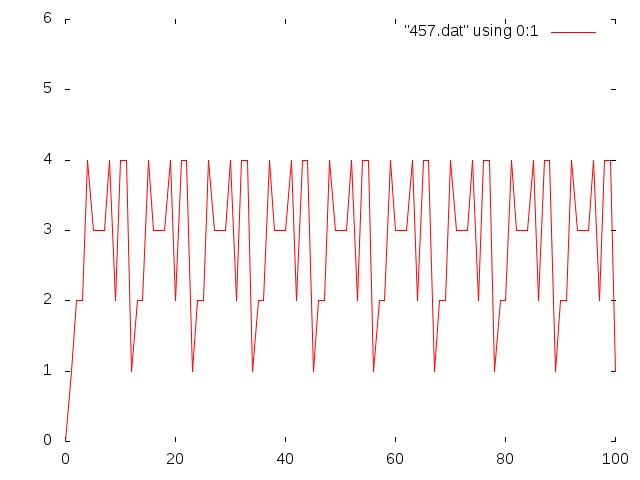}}\hfill{}\subfloat[Program Behavior from Sample 506]{\includegraphics[scale=0.25]{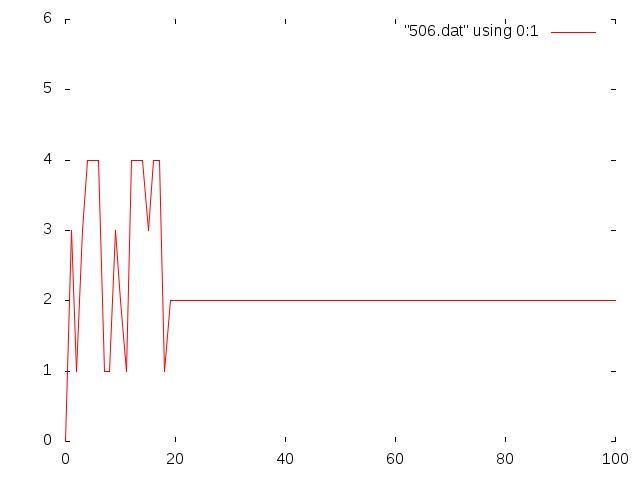}}

\subfloat[Program Behavior from Sample 1156]{

\includegraphics[scale=0.25]{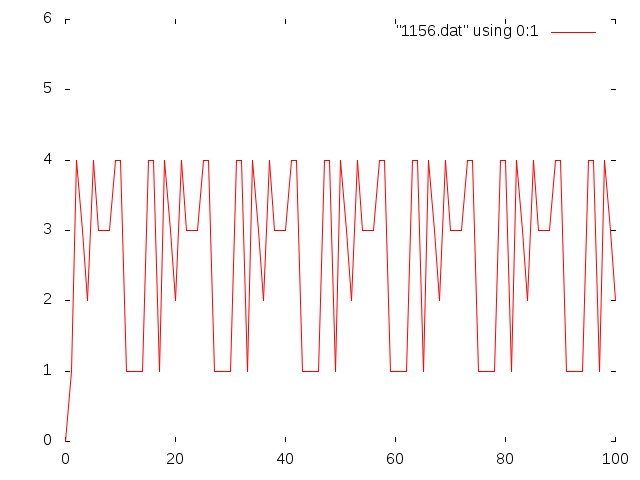}}\hfill{}\subfloat[Program Behavior from Sample 1519]{\includegraphics[scale=0.25]{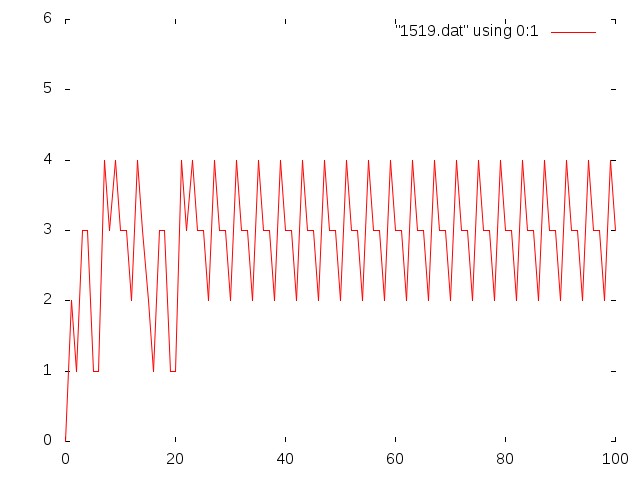}}\protect\caption{Four examples of asymptotically periodic code behaviors taken from
samples of code that reproduced continuously in experiment 2.\label{fig:behaviors}}
\end{figure}

Although the actual data summarized in table \ref{tab:Results:-Reproduction-vs.}
was one of the primary goals of this experiment, the operational characteristics
of the code itself were also of interest. The graphics in figure \ref{fig:behaviors}
represent some interesting code behaviors. In the graphs below, each
opcode was represented by a number in the following manner: START
= 0, COPY = 1, JUMP = 2, IF = 3, COND = 4, STOP = 5. With these numeric
values we can visualize program behavior graphically.

Although a myriad of behaviors were observed, in Figure 1 we have
selected some graphs of the different behaviors. Of specific interest
were the non-terminating programs. Each of these selections also exhibited
reproductive capabilities. Observe that although towards the beginning
of program execution, a variety of instructions were executed, eventually
the programs under consideration converged to some sort of periodic
behavior, a tendency we will call \emph{asymptotic periodicity}.

The graphs selected not only demonstrated asymptotic periodicity but
reproduced continuously in the sense that the progeny produced reached
the maximum allowable threshold.

\begin{table}
\centering{}\protect\caption{\label{tab:Results:-Reproduction-vs.}Results: Reproduction vs. Entropy}
\begin{tabular}{|c|c|c|}
\hline 
 & Alphabet 1 & Alphabet 2\tabularnewline
\hline 
\hline 
Total Simulations & 1,025,043 & 577,876\tabularnewline
\hline 
Average in Reproductions & 17.1433 & 39.3921\tabularnewline
\hline 
Std. Dev. in Reproductions & 23.7111 & 20.1516\tabularnewline
\hline 
Average in Entropy & 12.4556 & 34.8463\tabularnewline
\hline 
Std. Dev. in Entropy & 18.5525 & 17.6445\tabularnewline
\hline 
r-value & 0.870675 & 0.983243\tabularnewline
\hline 
\end{tabular}
\end{table}


\section{Applications to Biology\label{sec:Applications-to-Computional}}

Readers familiar with genetics will notice a strong correlation between
our computational formulations and genetics. In fact, genetics has
been an inspiration for this research, specifically with regard to
passive and hybrid forms of CTI. Recent research has been devoted
to the crossover between genetics and computer science. In fact, it
has been demonstrated in \cite{dna turing complete} that DNA computing
is Turing complete. This has grand implications for the very nature
of the definition of life itself.

The definition of life has always been somewhat controversial. In
\cite{dictionary}, life is defined as a characteristic that distinguishes
objects with self sustaining processes from those that do not. From
the vantage point of Biology, life is defined as a characteristic
of organisms that exhibit all or most of the following characteristics:
homeostasis, organization, metabolism, growth, adaptation, response
to stimuli, and reproduction \cite{bio char}. Many sources stress
the chemical nature of life, and mandate that living things contain
Carbon, Hydrogen, Nitrogen, Oxygen, Phosphorous and Sulfur \cite{life chemicals}.

In this paper, we introduce a definition for life as a self sustaining
passive or hybrid CTI process. Embedded in this definition are well
known assumptions regarding qualities one commonly associates with
living things. It is known that the purpose of life is to survive,
a fact that is redundant in the light of CTI. The fact that life is
fitness maximizing, and increases entropy follows directly from the
framework of CTI as well. Homeostasis follows naturally from the entropy
minimizing portion of the fundamental axiom of CTI. If an organism
is in homeostasis, this implies that the internal state space is ordered
and transitions can be predicted with high probability which implies
a lower entropy. Further, note the similarities with asymptotic periodicity
from the samples shown in \ref{fig:behaviors}. Of course, no discussion
about life would be complete without mentioning reproduction, which
we showed in this paper follows as a natural consequence of CTI.

As far as we know, Earth is the only planet in the universe known
to sustain life. However, recent advances in planetary sciences have
discovered the possibility of habitable zones in other solar systems.
This coupled with the observed robustness of life on Earth in extreme
environments, has raised interest the application of biological concepts
to the search for life beyond the confines of Earth. The vastness
of these environments forces us to rethink and generalize what we
know as life itself. One fantastic example of such an esoteric formulation
of life is the discovery of arsenic based life right here on Earth.
In December of 2010, NASA unveiled a discovery of a bacterium that
integrated Arsenic into its cellular structure and even its DNA. They
found that the bacterium was able to substitude arsenic for phosphorous,
one of the six essential ingredients of life. This challenges conventional
notions of life and seems to encourage a more general definition.
In this paper, we advocate a \emph{computational} approach to the
definition of life over a chemical one.

\section{Applications to Computer Science and Cyber Security\label{sec:Applications-to-Cyber}}

Applications to cyber security abound as well. In the past, computer
viruses typically served little more than the entertainment of the
author. Now, malicious softaware, or \emph{malware, }is a commoditized
industry. Patrons will hire teams of programmers to create specialized
malware for end goals such as espionage, ransome, or identity theft.
As such, mitigating these threats has become a priority in most industries
and even an interest in national security. For example, the US Defense
Advanced Research Projects Agency (DARPA) has issued a 'Cyber Grand
Challenge' offering up to \$2 million for designs of a fully automated
software defense system, capable of mitigating software exploitation
in real time \cite{darpa cyber grand challenge}. This is just one
example of increased efforts in this area. In fact, the US president
is proposing to increase the cyber defense budget to \$14 billion
in 2014, more than a billion dollar increase from that of the previous
year \cite{defense budget}.

How might this research be of merit in this field? Currently, most
anti-malware software (AM) stores hashes of actual malware samples
against which to compare new potential threats. This has multiple
disadvantages in that it ties users to a provider which may be an
issue if internet connectivity is a problem. Frequently such hashes
fail with polymorphic code.

Recently, much research has been conducted looking at a behavioral
approach to malware classification. While this is an exciting area
of resarch, approaches of this nature are typically burdened with
false positives rendering the solutions impractical.

This reserach offers the potential to view executable code in a new
light offering insights into two key areas of interests: the nature
of viruses, and the executable code itself, offering a new paradigm
to the software industry. Software serves a purpose. It has solid
attainable goals to suit the end user. Concurrently, it must function
within a specified range of parameters to preserve privacy, security,
and even safety. The effectiveness of a program under these guidelines
can be viewed as its fitness.

Regardless of whether we are analysing virus behavior or developing
code that optimizes its fitness, we need a reliable indicator as to
its behavior. One of the first successful defenses of malware was
to attack the very functions the AM product would call to detect it!
This is akin to asking the burgler if he is in your house. Thus behavior
detecting components must function at a 'lower' level than those to
which the malware may have access. Insuring this can be quite challenging.
Nevertheless, if such a framework were in place, the advantages have
enormous potential. In terms of viruses and payload detection, we
could apply analysis from section \ref{sub:Viruses}. Equivalently,
we could monitor program performance to scan for points of inadequacy.
Moreover, in the event of a breach or infection, we could employ the
methods of CTI as a means by which the code could heal itself. Again,
this necessitates a proper fitness function to evaluate the 'health
status' of the executable, and trusted behavioral data from the system.

Of course, there is a wide rift between the musings of theory and
the result of implementation, but this framework seems to provide
a promising direction for future efforts in this arena.

\section{Conclusion}

To recapitulate, in this paper, we extended the CTI framework presented
in \cite{cti1,cti2} to include genetic algorithms and Turing machines.
We defined classes of CTI including active, passive and hybrid. Finally,
we demonstrated that reproduction emerges as a consequence of the
axioms of CTI, both theoretically, and experimentally.

The concepts we presented have great potential for development in
future work. As far as the theoretical concepts of this paper are
concerned, we have a lot to do in terms of exploring the potential
of passive, active, and hybrid processes. It remains the stance of
this paper that hybrid CTI processes are present optimal solutions
but this remains to be shown. We would also like to further explore
simulation and programming using reproductive nested code. Further,
we would like to repeat the experiments of section \ref{sec:Experimentation}
with different instruction sets, and observe not only the behaviors,
but the phenotypes of the solutions in general, to observe overarching
patterns among diverse coding alphabets that acheive the same result.

In terms of multidisciplinary studies we have only scratched the surface
of the relationship between these concepts and computational molecular
biology, computer science, and cyber security.

Further, there is much to do in the way of advancing the theoretical
framework of CTI itself. In the next paper, we will visit the ramifications
of adding feedback into the intelligence process. Another paper will
focus on the global properties of intelligent agent.

\end{document}